\documentclass[10pt,twocolumn,letterpaper]{article}

\usepackage[pagenumbers]{iccv} 

\newcommand{\textitgray}[1]{\textit{\textcolor{gray}{#1}}}
\newcommand{\redtext}[1]{\textcolor[RGB]{205,38,38}{#1}}
\newcommand{\greentext}[1]{\textcolor[RGB]{34, 139, 34}{#1}}

\newcommand{\first}[1]{\textcolor[RGB]{141, 58, 148}{[#1]}}
\newcommand{\second}[1]{\textcolor[RGB]{195, 92, 36}{(#1)}}
\newcommand{\third}[1]{\textcolor[RGB]{82, 151, 199}{\underline{#1}}}

\usepackage{amsmath}
\usepackage{amssymb}
\usepackage{booktabs}
\usepackage{wrapfig}

\usepackage{pifont}
\usepackage{multirow}
\usepackage{multicol}
\usepackage{booktabs} 
\usepackage{rotating} 
\usepackage{makecell} 
\usepackage{tikz}

\usepackage[sectionbib]{chapterbib}

\definecolor{iccvblue}{rgb}{0.21,0.49,0.74}
\usepackage[pagebackref,breaklinks,colorlinks,allcolors=iccvblue]{hyperref}

\def\paperID{*}
\def\confName{ICCV}
\def\confYear{2025}

\title{VGGT-Long: Chunk it, Loop it, Align it, Pushing VGGT's Limits on Kilometer-scale Long RGB Sequences}

\author{
Kai Deng\textsuperscript{1},
Zexin Ti\textsuperscript{2},
Jiawei Xu\textsuperscript{1},
Jian Yang\textsuperscript{1,2},
Jin Xie\textsuperscript{2}\textsuperscript{$\dagger$} \\
\textsuperscript{1}College of Computer Science, Nankai University\\
\textsuperscript{2}School of Intelligence Science and Technology, Nanjing University\\
{\tt\small dengkai@mail.nankai.edu.cn, zexinti@smail.nju.edu.cn, jiaweixu@mail.nankai.edu.cn,}\\
{\tt\small csjyang@nankai.edu.cn, csjxie@nju.edu.cn}
}

\begin{document}

\twocolumn[{%
\renewcommand\twocolumn[1][]{#1}%
\maketitle

\begin{center}
  \centering
   \includegraphics[width=\linewidth]{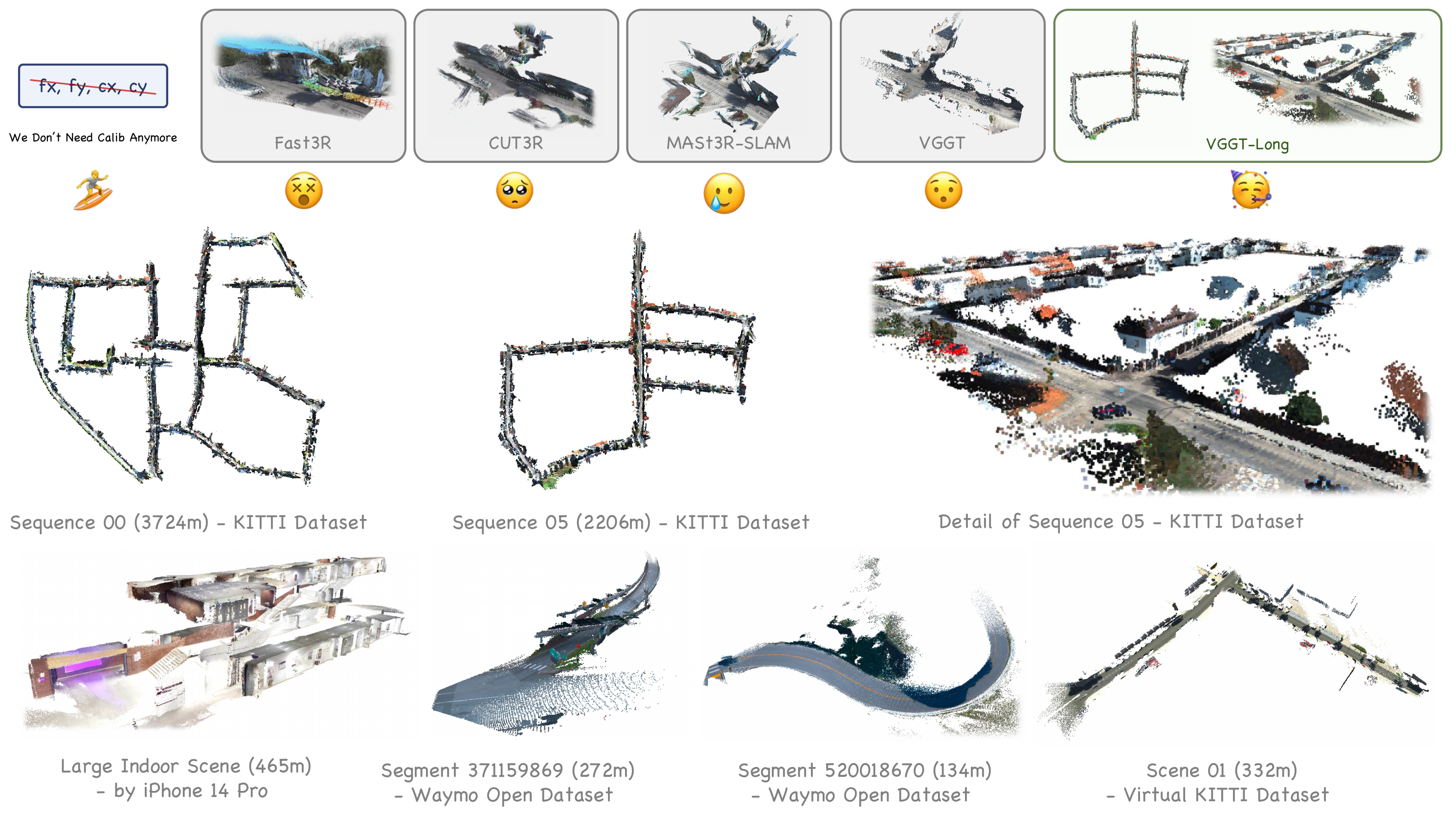}

   \captionof{figure}{
    For large-scale outdoor scenarios, previous work suffers from: 1) severe drift (CUT3R and Fast3R); 2) unable to complete the entire long sequence (MASt3R-SLAM and VGGT). Our method \textbf{VGGT-Long} is able to complete the reconstruction of the kilometer-scale scene while maintaining the accuracy of the scene.
   }
   \label{fig:overview}
\end{center}
}]

\footnotetext[1]{$^\dagger$  Corresponding author.}

\begin{abstract}
Foundation models for 3D vision have recently demonstrated remarkable capabilities in 3D perception. However, extending these models to large-scale RGB stream 3D reconstruction remains challenging due to memory limitations. In this work, we propose \textbf{VGGT-Long}, a simple yet effective system that pushes the limits of monocular 3D reconstruction to \textbf{kilometer-scale, unbounded outdoor environments}. Our approach addresses the scalability bottlenecks of existing models through a chunk-based processing strategy combined with overlapping alignment and lightweight loop closure optimization. Without requiring camera calibration, depth supervision or model retraining, VGGT-Long achieves trajectory and reconstruction performance comparable to traditional methods. We evaluate our method on KITTI, Waymo, and Virtual KITTI datasets. VGGT-Long not only runs successfully on long RGB sequences where foundation models typically fail, but also produces accurate and consistent geometry across various conditions. Our results highlight the potential of leveraging foundation models for scalable monocular 3D scene in real-world settings, especially for autonomous driving scenarios. Code is available at \url{https://github.com/DengKaiCQ/VGGT-Long}.
\end{abstract}    
\section{Introduction}
\label{sec:intro}

Perceiving 3D environments from monocular RGB streams is crucial for autonomous driving, yet existing methods struggle with kilometer-scale and uncalibrated sequences. Unlike the small-scale indoor 3D vision tasks, driving scenarios involve long trajectories with sparse frame correspondence, dynamic objects and challenging outdoor conditions. While some approaches \cite{dfvo, dvso, gigaslam} handle large-scale monocular scenes, they often depend on sophisticated multi-module pipelines or assume known camera intrinsics. Others leverage additional sensors (LiDAR \cite{zhang2014loam}, IMU \cite{zhang2015visual} or stereo  \cite{cvivsic2022soft2}), sidestepping the core challenge: scalable, calibration-free reconstruction from monocular RGB alone and it is a critical for autonomous systems.

A recent paradigm shift in 3D vision has witnessed the rise of end-to-end foundation models, largely based on the Transformer architecture \cite{vaswani2017attention}. A mainly of works from DUSt3R~\cite{dust3r} and MASt3R~\cite{mast3r} to CUT3R~\cite{cut3r}, Fast3R~\cite{fast3r}, and most recently VGGT~\cite{vggt}, aim to replace complex, multi-component SfM and SLAM pipelines with a single and unified deep learning model. These models are trained on massive datasets to integrate camera pose estimation, intrinsic parameter regression, and 3D scene representation (typically as a point map) into one cohesive framework. A key goal is to enable backpropagation of errors through the entire system, creating a powerful and versatile foundation model for 3D reconstruction that operates on raw and uncalibrated RGB inputs. However, foundation models like CUT3R and Fast3R still struggle with severe drift in outdoor environments, even on short sequences of a few dozen frames, which limits their practical applicability. In contrast, VGGT delivers remarkably stable and accurate local reconstructions, establishing it as the state-of-the-art in terms of reconstruction quality. Its primary limitation is not performance, but its immense computational and memory footprint.

The computational and memory demands of Transformer based foundation models severely limit their scalability. Standard self-attention \cite{vaswani2017attention} scales quadratically with input size, and while techniques like Flash-Attention \cite{dao2022flashattention, dao2023flashattention2} reduce compute complexity to linear. However, GPU memory requirement still remains prohibitive. For example, VGGT can just process 60 to 80 images on a 24 GiB RTX 4090 GPU scaling to a KITTI Seq 00 trajectory (about 4,600 frames) would require unreachable GPU memory requirement, far exceeding current hardware. This bottleneck confines such models to small-scale scenes, as both memory and drift accumulation become intractable over long sequences.


Our work is inspired by recent efforts to integrate foundation models into large-scale systems. A notable example is MASt3R-SLAM \cite{mast3r-slam}, which builds SLAM system on top of the MASt3R \cite{mast3r} model. To achieve global consistency, it employs pose graph optimization and bundle adjustment within its backend, which are standard components in modern SLAM systems.

This raises a fundamental question: must large-scale reconstruction always equate to system-level complexity? Our philosophy diverges significantly from this trend. We advocate for a minimalist approach that unlocks the inherent potential of the foundational model itself. We posit that VGGT is already a remarkably powerful engine for large-scale 3D perception, and the primary challenge is not a lack of capability, but a lack of scalability. Instead of building another full system around it, we ask: can we solve the problem with the minimal overhead?

To this end, we propose VGGT-Long, a framework that extends VGGT to long sequences through a simple yet effective framework that is processing the sequence in overlapping chunks, robustly aligning adjacent chunks, and correcting for drift using a high-quality loop closure module. This ``chunk-and-align'' paradigm avoids the need for a graph-based optimization backend (such as bundle adjustment \cite{triggs1999bundle}). It is a testament to the power of the underlying VGGT model, demonstrating that with the right strategy, its exceptional local reconstruction capabilities can be seamlessly stitched together to form a globally consistent, kilometer-scale map. Our work champions the idea that, \textbf{a sufficiently powerful base model may not necessarily require a system-level backend to assist}.

In summary, our contributions are as follows:

\begin{enumerate}
    \item We present the first system that successfully extends monocular 3D reconstruction models to kilometer-scale, unbounded outdoor scenes, without requiring camera calibration and depth supervision.
    
    \item We introduce a simple yet effective chunk-and-align pipeline that resolves the memory limitations of foundation models like VGGT on long video sequences, while achieving accuracy comparable to traditional methods with calibrated cameras.
    
    \item We address the accumulated Sim(3) drift problem inherent in processing long sequences with local models, demonstrating that VGGT can serve as a robust front-end for a large-scale reconstruction system without requiring a complex backend.

\end{enumerate}
\section{Related Work}
\label{sec:relatedwork}

\textbf{Structure-from-Motion (SfM)}. SfM methods estimate camera poses and sparse 3D structure from multi-view images. Classical SfM pipelines \cite{agarwal2011building, frahm2010building, schonberger2016structure} typically follow an incremental strategy: detecting keypoints \cite{detone2018superpoint,tyszkiewicz2020disk}, matching features \cite{chen2021learning, lindenberger2023lightglue} and refining poses through bundle adjustment \cite{triggs1999bundle}. While robust, these pipelines rely heavily on features extraction and are limited in textureless or ambiguous scenes. Recent deep learning methods aim to enhance or replace traditional modules. Hybrid frameworks such as PixSfM \cite{pixsfm} and DFSfM \cite{dfsfm} combine deep features with classical optimization to refine both tracks and structure. Fully differentiable SfM pipelines \cite{smith2024flowmap,tang2018ba, teed2018deepv2d, ummenhofer2017demon} further explore end-to-end learning of camera poses and depth but often suffer from scalability issues or poor generalization. VGGSfM \cite{vggsfm} demonstrates that learned systems can surpass classical SfM on real-world datasets by integrating dense features and multi-view consistency into a unified framework.

\textbf{SLAM and Visual Odometry}. To overcome the scalability issue of foundation models, researchers have explored to integrate them into larger SLAM systems. Traditional SLAM systems \cite{orbslam,orbslam2} rely on handcrafted features and optimization, while learning-based approaches \cite{gradslam,droidslam} integrate differentiable components into deep networks. However, these methods either scale poorly to long sequences or require pre-calibrated camera. MASt3R-SLAM~\cite{mast3r-slam} builds a sophisticated real-time SLAM framework around the MASt3R \cite{mast3r} without the calibration, employing complex backend machinery such as pose graph optimization and bundle adjustment to ensure global consistency. Concurrent work VGGT-SLAM \cite{vggt-slam} introduces a complete SLAM system that aligns submaps via SL(4)-based factor graph optimization for accurate indoor reconstruction. In contrast, our VGGT-Long targets large-scale outdoor scenes using a lightweight pipeline with Sim(3) transformations, prioritizing simplicity and scalability over a full SLAM framework.

\textbf{Transformer based 3D Vision Method}. A recent trend of 3D vision is the development of end-to-end foundation models, predominantly based on the Transformer architecture. These methods target dense geometry reconstruction from overlapping images, assuming unknown poses and unknown camera calibration. A pioneering line of work, including DUSt3R~\cite{dust3r} and its successor MASt3R~\cite{mast3r}, demonstrates the feasibility of jointly estimating camera parameters and dense 3D geometry from uncalibrated image pairs. Subsequent models like CUT3R~\cite{cut3r} and Fast3R~\cite{fast3r} further refined this paradigm. Most recently, VGGT~\cite{vggt} obtained a new state-of-the-art in reconstruction quality, producing remarkably stable and accurate local 3D maps from raw RGB inputs. However, a common limitation of these models is their significant computational and memory cost, which restricts their application to short image sequences.

Our work, VGGT-Long, distinguishes itself by adopting a minimalist philosophy. Instead of building a complex system, we focus on unlocking the full potential of the powerful VGGT model itself, using a simple yet effective chunk-and-align framework to extend its capabilities to long-sequence, large-scale scenarios with the minimal overhead.
\section{Method}

\begin{figure*}[t]
    \centering
    \includegraphics[width=\linewidth]{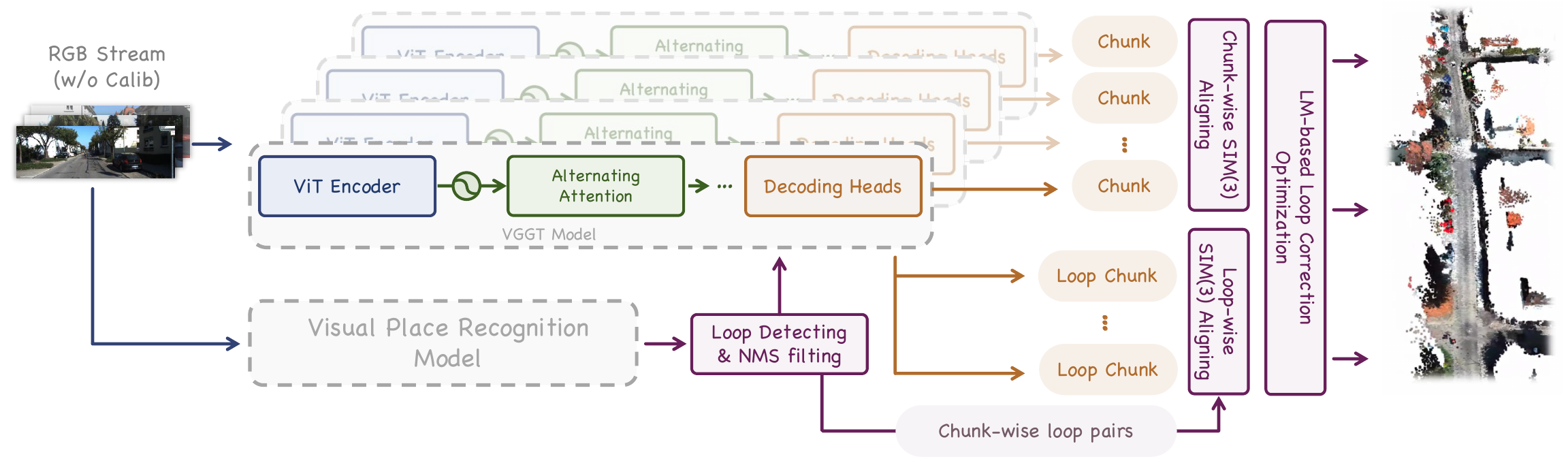}
    \caption{Overview of VGGT-Long. VGGT-Long processes long sequences by dividing them into different chunks, thereby handling the input RGB stream in a sliding window manner. We fully utilize VGGT's pointmap and confidence to perform lightweight loop closure and alignment on the output chunks, thus extending VGGT to long-sequence datasets for autonomous driving.}
    \label{fig:overview}
\end{figure*}

Our proposed method, VGGT-Long, processes long monocular RGB sequences by decomposing the problem into three stages: chunking, chunk-wise alignment, loop closure and loop correction. Our method maintains the local accuracy of VGGT while ensuring global accuracy when handling the outdoor long sequences.

\subsection{Sequence Chunking and Local Aligning with Confidence}
\label{sec:local_align}

\begin{figure}[t]
    \centering
    \includegraphics[width=\linewidth]{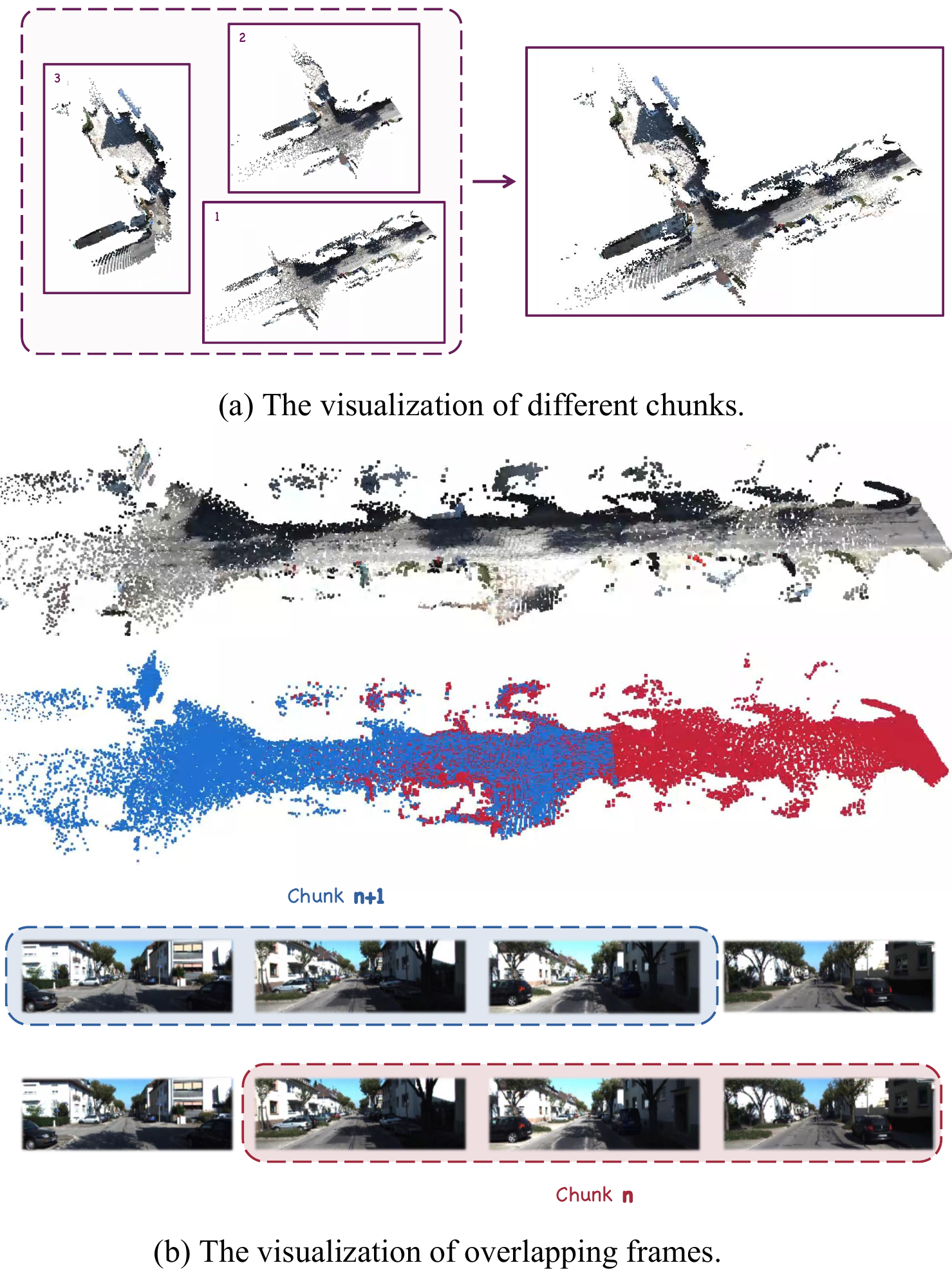}
    \caption{(a) VGGT-long divides a kilometer-scale sequence into different chunks for processing. (b) The alignments are derived from the consistency of overlapping frames in 3D space.}
    \label{fig:chunk_align}
\end{figure}

Given a long image sequence $ \mathcal{I} = \{\mathbf{I}_1, \dots, \mathbf{I}_N\} $, we partition it into $ K $ overlapping chunks. Let $L$ be the chunk size and $O$ be the overlap size. The $k$-th chunk, for $k=1, \dots, K$, is defined as the subsequence of frames indexed from $(k-1)(L-O)$ to $(k-1)(L-O) + L$. Each chunk $\mathcal{C}_k$ is processed independently by the VGGT model \cite{vggt}, which outputs a set of locally consistent camera poses and a 3D point map $\mathbf{P}_k\in \mathbb{R}^{H\times W\times 3}$ with the corresponding confidence values $\mathbf{c}_k\in \mathbb{R}^{H\times W}$. 

We fully leverage the confidence VGGT outputs to perform chunk alignment. In VGGT, the output confidence reflects the model's certainty about the point cloud. More specifically, in outdoor environments, objects such as the sky, oncoming vehicles, fast-moving pedestrians or raindrops during rainy conditions can affect the SIM(3) alignment calculation to some extent. VGGT's final output assigns lower confidence to these points, while assigning higher confidence to static scenes (such as buildings or stationary vehicles). Our alignment strategy is designed to achieve robust alignment based on this principle.

For each pair of adjacent chunks, $\mathcal{C}_k$ and $\mathcal{C}_{k+1}$, we identify a set of 3D point correspondences $\{(\mathbf{p}_k^i, \mathbf{p}_{k+1}^i)\}$ and $\{(\mathbf{c}_k^i, \mathbf{c}_{k+1}^i)\}$ within their overlapping region. To robustly estimate the relative Sim(3) transformation $\mathbf{S}_{k,k+1} \in \text{Sim}(3)$ that aligns $\mathcal{C}_{k+1}$ to $\mathcal{C}_k$, we employ an Iteratively Reweighted Least Squares (IRLS) optimization. The objective is to minimize the following robust cost function

\begin{equation}
\mathbf{S}_{k,k+1}^* = \arg\min_{\mathbf{S} \in \text{Sim}(3)} \sum_i \rho \left( \| \mathbf{p}_k^i - \mathbf{S} \mathbf{p}_{k+1}^i \|_2 \right),
\end{equation} 
where $\rho(\cdot)$ is the Huber loss function, which down-weights the influence of outliers. The IRLS procedure solves this non-linear problem by iteratively minimizing a weighted sum of squared errors

\begin{equation}
\mathbf{S}^{(t+1)} = \arg\min_{\mathbf{S} \in \text{Sim}(3)} \sum_i \mathbf{w}_i^{(t)} \| \mathbf{p}_k^i - \mathbf{S} \mathbf{p}_{k+1}^i \|_2^2.
\end{equation}

\begin{figure}[t]
    \centering
    \includegraphics[width=\linewidth]{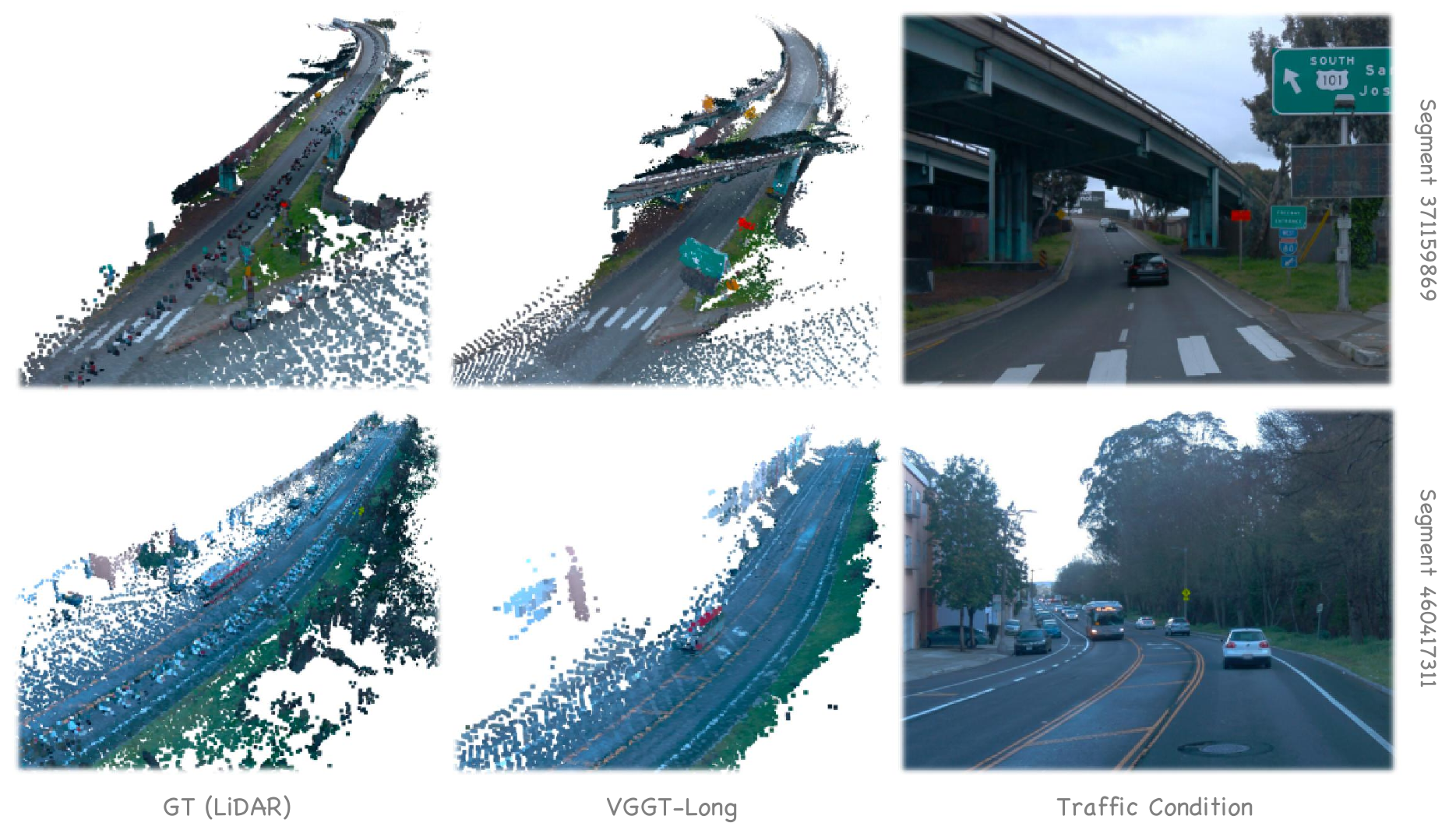}
    \caption{Confidence-aware alignment suppresses the influence of high-speed dynamic objects (such as vehicles) on alignment and reconstruction. It could be observed that higher-density vehicles cannot be effectively filtered out by the LiDAR, but VGGT-Long has the ability to handle this situation.}
    \label{fig:traffic}
\end{figure}

At each iteration $t$, the weight $\mathbf{w}_i^{(t)}$ for the $i$-th correspondence is a product of the model's confidence $\mathbf{c}_i$ and a robustness term derived from the Huber loss

\begin{equation}
\mathbf{w}_i^{(t)} = \mathbf{c}_i \cdot \frac{\rho'(r_i^{(t)})}{r_i^{(t)}},
\end{equation}
where $r_i^{(t)} = \| \mathbf{p}_k^i - \mathbf{S}^{(t)} \mathbf{p}_{k+1}^i \|_2$ is the residual from the previous iteration. Each weighted least-squares problem is solved efficiently in closed form using a weighted version of the Umeyama algorithm. Through this approach, we exclude low-confidence points (e.g., rapidly moving objects and sky regions) as outliers (see Fig. \ref{fig:traffic}). For medium-confidence objects (e.g., slowly moving vehicles), we assign reduced alignment weights, while concentrating higher weights on high-confidence structures (e.g., buildings). In our implementation, we directly discard points with confidence values below $0.1\times$ the median confidence of the entire chunk. The remaining low-confidence points that survive this filtering contribute minimally to the alignment process due to their attenuated weights.

\subsection{Loop Detection and Loop-wise SIM(3) Aligning}
\label{sec:loop_align}

To correct the accumulated drift inherent in sequential estimation, we perform loop closure detection across the entire sequence. This process involves identifying non-adjacent chunks as the same scene and robustly estimating the Sim(3) transformation between them.

First, we use a pre-trained Visual Place Recognition (VPR) model \cite{salad}, which leverages a DINOv2 backbone \cite{dinov2}, to extract a compact and descriptive global feature vector $ \mathbf{d}_i $ for each image $ \mathbf{I}_i $ in the sequence. These descriptors capture the high-level semantic and geometric content of the images.

With global descriptors for all images, we identify potential loop closure candidates. For each descriptor, we perform an efficient nearest neighbor search to find other images with high cosine similarity. A pair of images $(\mathbf{I}_i, \mathbf{I}_j)$ is considered as a potential loop closure if their similarity score exceeds a threshold $ \tau_s $ and their frame indices are sufficiently separated, i.e., $|i - j| > \Delta t_{min}$. To ensure that the detected loops are distinct and to avoid redundant matches in temporally close frames, we apply Non-Maximum Suppression (NMS). This filtering step selects the strongest match within a local time window, yielding a set of high-confidence image-level loop pairs.

For each validated loop pair $(\mathbf{I}_i, \mathbf{I}_j)$, we adopt a specialized strategy to generate a high-quality local reconstruction of the looped scene. We form a new, temporary image batch by concatenating sub-sequences of frames centered around the indices $i$ and $j$. This batch, containing temporally disjoint views of the same location, is then processed by the VGGT model. Unlike the sequential, sliding-window processing described in Sec. \ref{sec:local_align}, this approach provides VGGT with a more diverse, time-dispersed perspective, enabling a more robust reconstruction of the scene's geometry by leveraging a wider baseline.

\begin{figure}[t]
    \centering
    \includegraphics[width=\linewidth]{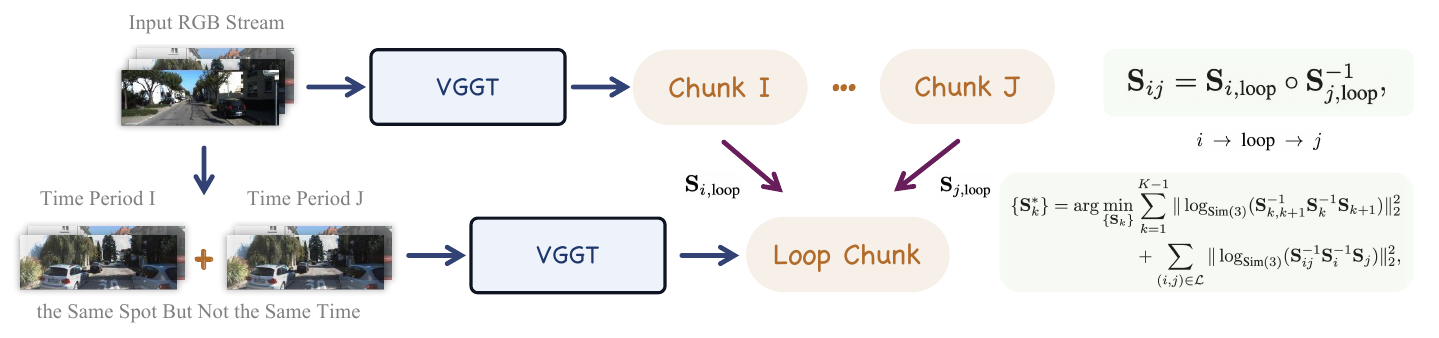}
    \caption{Loop-wise Sim(3) Alignment}
    \label{fig:loop_chunk}
\end{figure}

The resulting 3D point map, which we can call it the ``loop-centric'' chunk (see Figure \ref{fig:loop_chunk}), is then aligned with the original point maps of the corresponding chunks $\mathcal{C}_i$ and $\mathcal{C}_j$ (which were generated from temporally continuous frames). To compute the final loop-closing transformation $\mathbf{S}_{ij}$, we chain the alignments through this new chunk. Specifically, we compute the transformations from the original chunks to the loop-centric chunk and then compose them

\begin{equation}
\mathbf{S}_{ij} = \mathbf{S}_{i, \text{loop}} \circ \mathbf{S}_{j, \text{loop}}^{-1},
\end{equation}
where $\mathbf{S}_{i, \text{loop}}$ and $\mathbf{S}_{j, \text{loop}}$ are the transformations that align the loop-centric chunk to the coordinate frames of chunks $\mathcal{C}_i$ and $\mathcal{C}_j$ respectively (described in Sec. \ref{sec:local_align}). The composition $\mathbf{S}_{i, \text{loop}} \circ \mathbf{S}_{j, \text{loop}}^{-1}$ effectively chains the transformations to compute the direct alignment from chunk $\mathcal{C}_i$ to $\mathcal{C}_j$ (i.e., $i \rightarrow \text{loop} \rightarrow j$). This provides a robust geometric constraint for the global optimization by bridging the two distant chunks through a shared, high-quality local reconstruction.

\subsection{Global SIM(3) LM-based Optimization}

\begin{figure}[t]
    \centering
    \includegraphics[width=\linewidth]{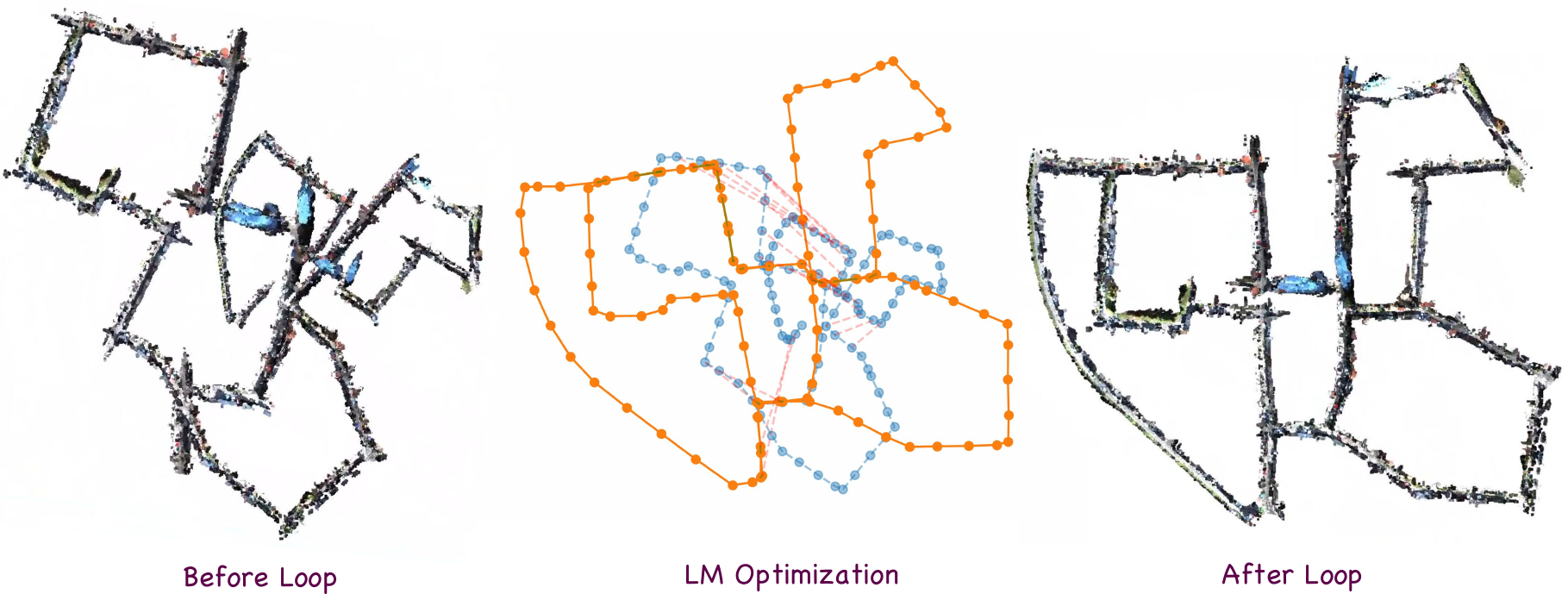}
    \caption{Without loop constraints, errors will be accumulated continuously at the kilometer scale. The use of Global LM Optimization can alleviate this accumulated error.}
    \label{fig:loop}
\end{figure}

\begin{figure}[t]
    \centering
    \includegraphics[width=0.7\linewidth]{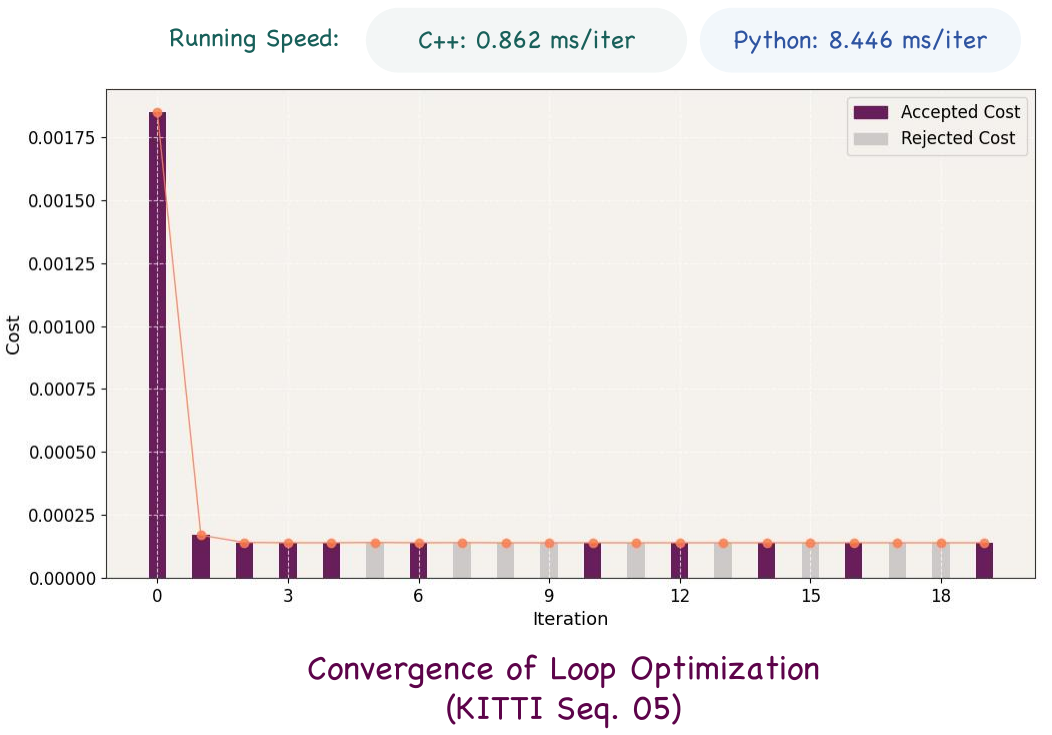}
    \caption{The loop optimization can be converged in just 3 iterations and can  achieve millisecond-level performance.}
    \label{fig:loop_convergence}
\end{figure}

To achieve global consistency, We follow \cite{strasdat2010scale, dpv-slam, gigaslam}, performing a global optimization over the Sim(3) transformations of all chunks. Instead of constructing a complex factor graph, we directly minimizes a non-linear least-squares objective function composed of two types of geometric constraints: sequential constraints from adjacent chunks (Sec. \ref{sec:local_align}) and loop closure constraints from non-adjacent chunks (Sec. \ref{sec:loop_align}). The goal is to jointly optimize the transformations $ \{\mathbf{S}_k\}_{k=1}^K $ for all chunks to be maximally consistent with these relative measurements.
The optimization problem is formulated as

\begin{equation}
\begin{split}
\{\mathbf{S}_k^*\} = \arg\min_{\{\mathbf{S}_k\}} & \sum_{k=1}^{K-1} \| \log_{\text{Sim}(3)}(\mathbf{S}_{k,k+1}^{-1} \mathbf{S}_k^{-1} \mathbf{S}_{k+1}) \|_2^2 \\
& + \sum_{(i,j) \in \mathcal{L}} \| \log_{\text{Sim}(3)}(\mathbf{S}_{ij}^{-1} \mathbf{S}_i^{-1} \mathbf{S}_j) \|_2^2
,\end{split}
\end{equation}
where $ \mathbf{S}_{k,k+1} $ is the relative transformation between adjacent chunks, $ \mathbf{S}_{ij} $ is the loop closure transformation between chunks $i$ and $j$, and $ \mathcal{L} $ is the set of all loop closures. The $ \log_{\text{Sim}(3)}(\cdot) $ map converts a Sim(3) transformation into its 7-dimensional tangent space representation in the Lie algebra $ \mathfrak{sim}(3) $, allowing for unconstrained optimization. We solve this non-linear least-squares problem efficiently using the Levenberg-Marquardt (LM) algorithm. The global optimization operates chunk-wise, maintaining a small set of SIM(3) variables (typically dozens even for KITTI, see Fig. \ref{fig:loop}). Convergence is achieved within few iterations, with processing times in milliseconds (see Fig. \ref{fig:loop_convergence}).

\section{Experiments}

\begin{table*}[htbp]
  \centering
  \label{tab:table1}
  \centering
 \resizebox{\linewidth}{!}{\textbf{}
    \begin{tabular}{c|l|ccc|c|c|ccccccccccc}
    \toprule
        &\textbf{Methods} & \textbf{LC} & \textbf{Calibration} & \textbf{Recon.} & \textbf{Avg.} & \textbf{Avg.$^*$} & \textbf{00} & \textbf{01} & \textbf{02} & \textbf{03} & \textbf{04} & \textbf{05} & \textbf{06} & \textbf{07} & \textbf{08} & \textbf{09} & \textbf{10}   \\ \midrule

        &\textitgray{seq. frames} & \textitgray{-} & \textitgray{-}&\textitgray{-}&	\textitgray{2109}&	\textitgray{2210} &	\textitgray{4542}&	\textitgray{1101}&	\textitgray{4661}&	\textitgray{801}&	\textitgray{271}&	\textitgray{2761}&	\textitgray{1101}&	\textitgray{1101}&	\textitgray{4071}&	\textitgray{1591}&	\textitgray{1201}\\
        
        &\textitgray{seq. length (m)} & \textitgray{-} & \textitgray{-}& \textitgray{-} & \textitgray{2012.243} & \textitgray{1968.147} & \textitgray{3724.19} & \textitgray{2453.20} & \textitgray{5067.23} & \textitgray{560.89} & \textitgray{393.65} & \textitgray{2205.58} & \textitgray{1232.88} & \textitgray{649.70} & \textitgray{3222.80} & \textitgray{1705.05} & \textitgray{919.52} \\
         	
        &\textitgray{seq. speed (m / frame)} & \textitgray{-} & \textitgray{-}& \textitgray{-} & \textitgray{0.95} & \textitgray{0.89} & \textitgray{0.82 } & \textitgray{\underline{2.23} } & \textitgray{1.09 } & \textitgray{0.70} & \textitgray{1.45 } & \textitgray{0.80} & \textitgray{1.12} & \textitgray{0.59} & \textitgray{0.79} & \textitgray{1.07} & \textitgray{0.77} \\
        
        &\textitgray{contains loop} & \textitgray{-} & \textitgray{-} & \textitgray{-}& \textitgray{-} & \textitgray{-} & \textitgray{\ding{51}} & \textitgray{\ding{55}} & \textitgray{\ding{51}} & \textitgray{\ding{55}} & \textitgray{\ding{55}} & \textitgray{\ding{51}} & \textitgray{\ding{51}} & \textitgray{\ding{51}} & \textitgray{\ding{55}} & \textitgray{\ding{51}} & \textitgray{\ding{55}} \\
        
        \midrule
        \multirow{3}{*}{\begin{turn}{90}Classic\end{turn}}&ORB-SLAM2 (w/o LC) \cite{orbslam2} & \redtext{\ding{55}} & \redtext{\textit{Required}}& \redtext{\textit{Sparse}}  & 69.727 & 26.480 & 40.65  & 502.20  & 47.82  & \first{0.94}  & 1.30  & 29.95  & 40.82  & 16.04  & \second{43.09}  & 38.77  & \first{5.42}   \\ 
        
        & ORB-SLAM2 (w/ LC) \cite{orbslam2} & \greentext{\ding{51}} & \redtext{\textit{Required}}& \redtext{\textit{Sparse}}  & 54.816 & \first{9.464} & \first{6.03}  & 508.34  & \first{14.76}  & \second{1.02}  & 1.57  & \first{4.04}  & 11.16  & \second{2.19}  & \first{38.85}  & \first{8.39}  & \second{6.63}   \\ 
        
        & LDSO \cite{ldso} & \greentext{\ding{51}} & \redtext{\textit{Required}}& \redtext{\textit{Sparse}}  & \first{22.425} & 23.500  & 9.32  & \second{11.68}  & \second{31.98}  & 2.85  & 1.22  & \second{5.10}  & 13.55  & 2.96  & 129.02  & \third{21.64}  & 17.36   \\ 

        \midrule
        
        \multirow{13}{*}{\begin{turn}{90}Learning Based\end{turn}}&DROID-VO \cite{droidslam} & \redtext{\ding{55}} & \redtext{\textit{Required}}& \greentext{\textit{Dense}}  & 54.188 & 51.187 & 98.43  & 84.20  & 108.80  & 2.58  & 0.93  & 59.27  & 64.40  & 24.20  & 64.55  & 71.80  & 16.91   \\ 
        
        &DPVO \cite{dpvo}& \redtext{\ding{55}} & \redtext{\textit{Required}}& \redtext{\textit{Sparse}}  & 53.609 & 57.701 & 113.21  & 12.69  & 123.40  & \third{2.09}  & \first{0.68}  & 58.96  & 54.78  & 19.26  & 115.90  & 75.10  & 13.63     \\ 
        
        &DROID-SLAM \cite{droidslam} & - & \redtext{\textit{Required}}& \greentext{\textit{Dense}}  & 100.278 & 75.846  & 92.10  & 344.60  & 107.61 & 2.38  & 1.00  & 118.50  & 62.47  & 21.78  & 161.60  & 72.32 & 118.70   \\ 
        
        &DPV-SLAM \cite{dpv-slam} & \greentext{\ding{51}} & \redtext{\textit{Required}}& \redtext{\textit{Sparse}}  & 53.034 & 57.187  & 112.80  & \first{11.50}  & 123.53  & 2.50  & \third{0.81}  & 57.80  & 54.86  & 18.77  & 110.49  & 76.66  & \third{13.65}   \\
        
        &DPV-SLAM++ \cite{dpv-slam} & \greentext{\ding{51}} & \redtext{\textit{Required}}& \redtext{\textit{Sparse}}  & 25.749 & 27.138 & \third{8.30}  & \third{11.86}  & 39.64  & 2.50  & \second{0.78}  & \third{5.74}  & 11.60  & \first{1.52}  & 110.90  & 76.70  & 13.70    \\ 

        \cmidrule(lr){2-18}
        
        
        &MASt3R-SLAM \cite{mast3r-slam} & \greentext{\ding{51}} & \greentext{\textit{No Need}}& \greentext{\textit{Dense}} &	/ & /  & \textitgray{TL} &	\textitgray{\textitgray{TL}} & \textitgray{TL} & \textitgray{TL} & \textitgray{TL} & \textitgray{TL} & \textitgray{TL}  & \textitgray{TL} & \textitgray{TL} & \textitgray{TL} & \textitgray{TL}  \\ 
        &CUT3R \cite{cut3r} & \redtext{\ding{55}} & \greentext{\textit{No Need}}& \greentext{\textit{Dense}} &	/ & /  & \textitgray{OOM} &	\textitgray{OOM} & \textitgray{OOM} & 148.07 & 22.31 & \textitgray{OOM} & \textitgray{OOM}  & \textitgray{OOM} & \textitgray{OOM} & \textitgray{OOM} & \textitgray{OOM}  \\ 
        &Fast3R \cite{fast3r} & \redtext{\ding{55}} & \greentext{\textit{No Need}}& \greentext{\textit{Dense}} &	/ & /  & \textitgray{OOM} &	\textitgray{OOM} & \textitgray{OOM} & \textitgray{OOM} & \textitgray{OOM} & \textitgray{OOM} & \textitgray{OOM}  & \textitgray{OOM} & \textitgray{OOM} & \textitgray{OOM} & \textitgray{OOM}  \\ 
        
        &VGGT \cite{vggt} & \redtext{\ding{55}} & \greentext{\textit{No Need}}& \greentext{\textit{Dense}} &	/ & /  & \textitgray{OOM} &	\textitgray{OOM} & \textitgray{OOM} & \textitgray{OOM} & \textitgray{OOM} & \textitgray{OOM} & \textitgray{OOM}  & \textitgray{OOM} & \textitgray{OOM} & \textitgray{OOM} & \textitgray{OOM}  \\ 

        \cmidrule(lr){2-18}
        
        &\textbf{VGGT-Long} (Chunk Size=30) & \greentext{\ding{51}} & \greentext{\textit{No Need}}& \greentext{\textit{Dense}} &	44.713 &	39.564&   9.06& 	96.20& 	99.95& 	19.76& 	11.36& 	10.20& 	\third{10.04}& 	4.00& 	139.79& 	50.53& 	40.94   \\ 
        &\textbf{VGGT-Long} (Chunk Size=60) & \greentext{\ding{51}} & \greentext{\textit{No Need}}& \greentext{\textit{Dense}} &	26.358 &	\second{19.298}&   \second{8.06}& 	96.96& 	\third{34.16}& 	6.83& 	4.16& 	9.15& 	\first{4.68}& 	\second{2.68}& 	\third{63.15}& 	\third{32.24}& 	27.87   \\ 
        &\textbf{VGGT-Long} (Chunk Size=90) & \greentext{\ding{51}} & \greentext{\textit{No Need}}& \greentext{\textit{Dense}} &	\second{22.718} &	\third{20.938}&   11.97& 	40.51& 	49.85& 	5.41& 	2.86 &	9.88& 	\second{6.07}& 	3.47& 	66.27& 	32.27& 	21.33   \\ 
        &\textbf{VGGT-Long} (Chunk Size=120) & \greentext{\ding{51}} & \greentext{\textit{No Need}}& \greentext{\textit{Dense}} &	\third{25.597} &	22.814& 16.13& 	53.43& 	51.98& 	4.37& 	2.15& 	12.69& 	11.33& 	3.603& 	70.29& 	34.55& 	21.05   \\ 
        \bottomrule
        
    \end{tabular}
  }
    \caption{Camera Tracking Results (ATE RMSE [m] $\downarrow$) on the KITTI Dataset Odometry Track. The overlap was set to half of the chunk size. \textitgray{[LC]} denotes \textit{loop closure}. Since the Seq. 01 is a high-speed sequence, its movement pattern is significantly different from those of other sequences. \textitgray{[Avg.$^*$]} shows the mean ATE excluding Seq 01. \textit{VGGT-Long} achieves relatively good tracking accuracy without the input of camera intrinsic matrix. \textitgray{[OOM]} is short for \textit{CUDA Out-Of-Memory} on a single RTX 4090. \textitgray{[TL]} is short for \textit{Tracking Lost}. See Fig. \ref{fig:kitti_traj} for trajectory visual results. Color definition: \first{first}, \second{second}, \third{third}.}
  \label{table:kitti_ate}

\end{table*}

\begin{table*}[h]
  \centering
 \resizebox{0.95\linewidth}{!}{\textbf{}
\begin{tabular}{l|c|c|ccccccccc}
\toprule
\textbf{Segment ID}       & \textbf{Calib.}   & \textbf{Avg.} & \textbf{163453191} & \textbf{183829460} & \textbf{315615587} & \textbf{346181117} & \textbf{371159869} & \textbf{405841035} & \textbf{460417311} & \textbf{520018670} & \textbf{610454533} \\
\midrule
\textitgray{Frame num.}        & \textitgray{-}        & \textitgray{198}           & \textitgray{198}                & \textitgray{199}                & \textitgray{199}                & \textitgray{199}                & \textitgray{196}                & \textitgray{199}                & \textitgray{198}                & \textitgray{199}                & \textitgray{198}                \\
\textitgray{Segment length}   & \textitgray{-}        & \textitgray{172.533}       & \textitgray{159.963}            & \textitgray{42.301}             & \textitgray{165.149}            & \textitgray{351.213}            & \textitgray{272.661}            & \textitgray{85.743}             & \textitgray{265.906}            & \textitgray{134.552}            & \textitgray{62.739}             \\
\textitgray{Segment speed}    & \textitgray{-}        & \textitgray{0.871}         & \textitgray{0.808}              & \textitgray{0.213}              & \textitgray{0.830}              & \textitgray{1.765}              & \textitgray{1.391}              & \textitgray{0.431}              & \textitgray{1.343}             & \textitgray{0.676}              & \textitgray{0.317}              \\
\textitgray{Traffic}          & \textitgray{-}        & \textitgray{-}             & \textitgray{Low}                & \textitgray{High}               & \textitgray{Low}                & \textitgray{Low}                & \textitgray{Medium}                & \textitgray{Low}                & \textitgray{Medium}             & \textitgray{Low}                & \textitgray{High}               \\
\midrule
DROID SLAM \cite{droidslam}      & \redtext{\textit{Required}} & \second{4.396}         & \second{3.705}              & \first{0.301}              & \first{0.447}              & \second{8.653}              & 9.320              & 7.621              & \second{4.170}              & \textitgray{TL}                 & \first{0.264}              \\
MASt3R-SLAM \cite{mast3r-slam}     & \greentext{\textit{No Need}}  & 5.560         & 4.500              & \second{0.556}              & 1.833              & 12.544             & \second{8.601}              & \first{1.412}              & 5.428              & \second{7.910}              & 1.195              \\
CUT3R \cite{cut3r}           & \greentext{\textit{No Need}}  & 9.872         & 8.781              & 3.810              & 5.790              & 24.015             & 13.070             & 7.261              & 13.206             & 8.597              & 3.229              \\
Fast3R \cite{fast3r}           & \greentext{\textit{No Need}}  & /             & \textitgray{OOM}                & \textitgray{OOM}                & \textitgray{OOM}                & \textitgray{OOM}                & \textitgray{OOM}                & \textitgray{OOM}                & \textitgray{OOM}                & \textitgray{OOM}                & \textitgray{OOM}                \\
VGGT \cite{vggt}            & \greentext{\textit{No Need}}  & /             & \textitgray{OOM}                & \textitgray{OOM}                & \textitgray{OOM}                & \textitgray{OOM}                & \textitgray{OOM}                & \textitgray{OOM}                & \textitgray{OOM}                & \textitgray{OOM}                & \textitgray{OOM}                \\
\midrule
\textbf{VGGT-Long (Ours)} & \greentext{\textit{No Need}}  & \first{1.996}         & \first{1.753}              & 2.629              & \second{0.559}              & \first{3.452}              & \first{3.343}              & \second{1.444}              & \first{1.541}              & \first{2.547}              & \second{0.455}      \\
\bottomrule
\end{tabular}
  }
  \caption{Camera Tracking Results (ATE RMSE [m] $\downarrow$) on the Waymo Open Dataset. Color definition: \first{first}, \second{second}.}
  \label{table:waymo_ate}
\end{table*}

\begin{table*}[h]
  \centering
 \resizebox{0.9\linewidth}{!}{\textbf{}
\begin{tabular}{l|l|c|c|ccccccccc}
\toprule
\textbf{Segment ID}  & \textbf{Metric} & \textbf{Calib.}      & \textbf{Avg.} & \textbf{163453191} & \textbf{183829460} & \textbf{315615587} & \textbf{346181117} & \textbf{371159869} & \textbf{405841035} & \textbf{460417311} & \textbf{520018670} & \textbf{610454533} \\
\midrule
\textitgray{Frame num.}            & \textitgray{-}               & \textitgray{-}                    & \textitgray{198}           & \textitgray{198}                & \textitgray{199}                & \textitgray{199}                & \textitgray{199}                & \textitgray{196 }               & \textitgray{199}                & \textitgray{198}                & \textitgray{199}                & \textitgray{198}                \\
\textitgray{Segment length}       & \textitgray{-}               & \textitgray{-}                    & \textitgray{172.533}       & \textitgray{159.963}            & \textitgray{42.301}             & \textitgray{165.149}            & \textitgray{351.213}            & \textitgray{272.661}            & \textitgray{85.743}             & \textitgray{265.906}            & \textitgray{134.552}            & \textitgray{62.739}             \\
\textitgray{Segment speed}        & \textitgray{-}               & \textitgray{-}                    & \textitgray{0.871}         & \textitgray{0.808}              & \textitgray{0.213}              & \textitgray{0.830}              & \textitgray{1.765}              & \textitgray{1.391}              & \textitgray{0.431}              & \textitgray{1.343}              & \textitgray{0.676}              & \textitgray{0.317}              \\
\textitgray{Traffic}              & \textitgray{- }              & \textitgray{-}                    & \textitgray{-}             & \textitgray{Low}                & \textitgray{High}               & \textitgray{Low}                & \textitgray{Low}                & \textitgray{Medium}                & \textitgray{Low}                & \textitgray{Medium}             & \textitgray{Low}                & \textitgray{High}               \\
\midrule

\multirow{3}{*}{DROID-SLAM \cite{droidslam}} & Accuracy $\downarrow$           & \multirow{3}{*}{\redtext{\textit{Required}}}             & \textitgray{1.201}         & \first{0.781}              & 1.136              & 2.247              & 2.393              & \first{1.090}              & \first{0.539}              & \first{0.740}              & \textitgray{TL}                 & \first{0.677}              \\
 & Completeness $\downarrow$           &  & \textitgray{8.540}         & 4.610              & 10.245             & 5.540              & 8.669              & 8.592              & 11.144             & 5.320              & \textitgray{TL}                 & 14.201             \\
 & Chamfer $\downarrow$        &  & \textitgray{4.870}         & 2.696              & 5.691              & 3.893              & 5.531              & 4.841              & 5.842              & 3.030              & \textitgray{TL}                 & 7.439              \\
 \midrule
\multirow{3}{*}{MASt3R-SLAM \cite{mast3r-slam}}          & Accuracy $\downarrow$           & \multirow{3}{*}{\greentext{\textit{No Need}}}              & 3.772         & 3.189              & 2.988              & 3.787              & 4.689              & 4.436              & 1.166              & 4.637              & 6.417              & 2.637              \\
 & Completeness $\downarrow$          &  & 3.177         & \first{1.715}              & \first{3.284}              & 2.047              & \first{2.981}              & \first{2.679}              & \first{2.895}              & 2.002              & 4.429              & 6.560              \\
 & Chamfer $\downarrow$        &                      & 3.474         & 2.452              & 3.136              & 2.917              & 3.835              & 3.558              & 2.031              & 3.319              & 5.423              & 4.599              \\
 \midrule
\multirow{3}{*}{CUT3R \cite{cut3r}}                & Accuracy $\downarrow$           & \multirow{3}{*}{\greentext{\textit{No Need}}}              & 3.884         & 3.580              & 1.144              & 2.418              & 3.712              & 3.679              & 4.346              & 2.012     & 12.320             & 1.744              \\
                     & Completeness $\downarrow$          &  & 6.801         & 8.251              & 9.352              & 8.748              & 8.537              & 5.467              & 3.393              & 6.164     & \first{2.302}              & 8.999              \\
                     & Chamfer $\downarrow$        &  & 5.343         & 5.916              & 5.248              & 5.583              & 6.125              & 4.573              & 3.869              & 4.088     & 7.311              & 5.371              \\
                     \midrule
\multirow{3}{*}{\textbf{VGGT-Long (Ours)}}     & Accuracy $\downarrow$           & \multirow{3}{*}{\greentext{\textit{No Need}}}              & \first{1.182}         & 1.002     & \first{0.395}     & \first{0.925}              & \first{1.668}              & 2.580              & 0.679              & 0.784              & \first{1.358}              & 1.246              \\
 & Completeness $\downarrow$          &  & \first{2.860}         & 2.762     & 3.417     & \first{1.738}              & 3.261              & 2.791              & 3.216              & \first{1.840}              & 4.694              & \first{2.022}              \\
 & Chamfer $\downarrow$        &  & \first{2.021}         & \first{1.882}     & \first{1.906}     & \first{1.331}              & \first{2.465}              & \first{2.685}              & \first{1.948}              & \first{1.312}              & \first{3.026}              & \first{1.634}             \\
\bottomrule
\end{tabular}
  }
  \caption{Point Map Estimation Results on the Waymo Open Dataset. The GT point cloud is collected by LiDAR, which on vehicles is typically scanning at a height below human eye level, the GT point cloud's coverage is actually narrower than the observation range of RGB cameras. As can be seen in \textitgray{Segment 371159869}, the LiDAR-captured GT point cloud fails to perceive the 3D structure of the overpass, whereas almost all visual-based methods can, to varying degrees, detect and reconstruct this structure. Therefore, the numerical values in this table should be taken as reference only, and readers are encouraged to refer to Fig. \ref{fig:visual_cmp_1} to Fig. \ref{fig:visual_cmp_4} for a more intuitive visual comparison between methods. However, whether by visual comparison or the reference numerical values, it is evident that VGGT-Long achieves high reconstruction performance than other methods. Color definition: \first{first}.}
  \label{table:waymo_point}
\end{table*}

\begin{table*}[h]
  \centering
 \resizebox{0.9\linewidth}{!}{\textbf{}

\begin{tabular}{l|c|cccccc|cccccc}
\toprule
 & \textbf{Calib.} & \multicolumn{6}{c}{\textbf{Scene 01} ~~~\textitgray{Speed: 0.743, Num.: 447, Length 332.487 m}} & \multicolumn{6}{c}{\textbf{Scene 02} ~~~\textitgray{Speed: 0.509, Num.: 223, Length 113.603 m}} \\
\midrule
\textitgray{Condition}                     & -               & \textitgray{Clone}        & \textitgray{Fog}         & \textitgray{Morning}          & \textitgray{Overcast}        & \textitgray{Rain}       & \textitgray{Sunset}      & \textitgray{Clone}        & \textitgray{Fog}         & \textitgray{Morning}          & \textitgray{Overcast}        & \textitgray{Rain}       & \textitgray{Sunset}      \\
\midrule
DROID-SLAM \cite{droidslam}                    & \redtext{\textit{Required}}        & \second{1.0267}       & \second{1.8679}      & \second{0.9886}           & \second{1.0146}          & \first{0.7763}     & \first{1.1453}      & \first{0.0981}      & \first{0.0401}      & \first{0.0493}              & \first{0.0478}       & \first{0.0363}      & \first{0.1127}      \\
MASt3R-SLAM \cite{mast3r-slam}                  & \greentext{\textit{No Need}}         & \textitgray{TL}           & \textitgray{TL}          & \textitgray{TL}               & \textitgray{TL}              & \textitgray{TL}         & \textitgray{TL}          & \textitgray{TL}          & \textitgray{TL}          & \textitgray{TL}                  & \textitgray{TL}           & \textitgray{TL}          & \textitgray{TL}          \\
CUT3R \cite{cut3r}                        & \greentext{\textit{No Need}}         & 43.3039      & 62.1908     & 50.6081          & 38.7345         & 51.5482    & 43.7854     & 23.7711     & 9.9484      & 28.4150             & 24.6440      & 7.9632      & 25.9732     \\
Fast3R \cite{fast3r}                       & \greentext{\textit{No Need}}         & \textitgray{OOM}          & \textitgray{OOM}         & \textitgray{OOM}              & \textitgray{OOM}             & \textitgray{OOM}        & \textitgray{OOM}         & \textitgray{OOM}         & \textitgray{OOM}         & \textitgray{OOM}                 & \textitgray{OOM}          & \textitgray{OOM}         & \textitgray{OOM}         \\
VGGT \cite{vggt}                         & \greentext{\textit{No Need}}         & \textitgray{OOM}          & \textitgray{OOM}         & \textitgray{OOM}              & \textitgray{OOM}             & \textitgray{OOM}        & \textitgray{OOM}         & \textitgray{OOM}         & \textitgray{OOM}         & \textitgray{OOM}                 & \textitgray{OOM}          & \textitgray{OOM}         & \textitgray{OOM}         \\
\textbf{VGGT-Long (Ours)}              & \greentext{\textit{No Need}}         & \first{0.7632}       & \first{0.8739}      & \first{0.9279}           & \first{0.6704}          & \second{1.7991}     & \second{1.2593}      & \second{0.7229}      & \second{0.7086}      & \second{0.7214}              & \second{0.6809}       & \second{0.6930}      & \second{0.6890}      \\
\midrule
                              & \textbf{Calib.}          & \multicolumn{6}{c}{\textbf{Scene 06} ~~~ \textitgray{Speed: 0.192, Num.: 270, Length 51.899 m}}           & \multicolumn{6}{c}{\textbf{Scene 18} ~~~\textitgray{Speed: 0.751, Num.: 339, Length 254.443 m} }          \\
                              \midrule
\textitgray{Condition}                      & -               & \textitgray{Clone}        & \textitgray{Fog}         & \textitgray{Morning}          & \textitgray{Overcast}        & \textitgray{Rain}       & \textitgray{Sunset}      & \textitgray{Clone}        & \textitgray{Fog}         & \textitgray{Morning}          & \textitgray{Overcast}        & \textitgray{Rain}       & \textitgray{Sunset}     \\
\midrule
DROID-SLAM \cite{droidslam}                   & \redtext{\textit{Required}}        & \first{0.0633}       & \first{0.0243}      & \first{0.0302}           & \first{0.0509}          & \textitgray{TL}         & \first{0.0197}      & \first{2.4775}      & \second{2.0316}      & \second{1.8939}              & \second{2.3317}       & \second{2.5497}      & \second{1.9434}      \\
MASt3R-SLAM \cite{mast3r-slam}                  & \greentext{\textit{No Need}}         & \textitgray{TL}           & \textitgray{TL}          & \textitgray{TL}               & \textitgray{TL}              & \textitgray{TL}         & \textitgray{TL}          & \textitgray{TL}          & \textitgray{TL}          & \textit{\textitgray{TL}}         & \textitgray{TL}           & \textitgray{TL}          & \textitgray{TL}          \\
CUT3R \cite{cut3r}                        & \greentext{\textit{No Need}}         & 0.8358       & \second{0.4078}      & 0.5990           & 0.7204          & \second{1.0590}     & 1.0127      & 19.4402     & 8.6279      & \textit{6.7201}     & 20.2124      & 16.7768     & 31.1191     \\
Fast3R \cite{fast3r}                       & \greentext{\textit{No Need}}         & \textitgray{OOM}          & \textitgray{OOM}         & \textitgray{OOM}              & \textitgray{OOM}             & \textitgray{OOM}        & \textitgray{OOM}         & \textitgray{OOM}         & \textitgray{OOM}         & \textit{\textitgray{OOM}}        & \textitgray{OOM}          & \textitgray{OOM}         & \textitgray{OOM}         \\
VGGT \cite{vggt}                         & \greentext{\textit{No Need}}         & \textitgray{OOM}          & \textitgray{OOM}         & \textit{\textitgray{OOM}}     & \textit{\textitgray{OOM}}    & \textitgray{OOM}        & \textitgray{OOM}         & \textitgray{OOM}         & \textitgray{OOM}         & \textitgray{OOM}                 & \textitgray{OOM}          & \textitgray{OOM}         & \textitgray{OOM}         \\
\textbf{VGGT-Long (Ours)}              & \greentext{\textit{No Need}}         & \second{0.3649}       & 0.5425      & \second{0.3761}           & \second{0.4024}          & \first{0.5585}     & \second{0.3819}      & \second{1.6509}      & \first{0.7969}      & \first{1.2879}              & \first{1.2556}       & \first{1.6480}      & \first{1.7402}      \\
\midrule
                              & \textbf{Calib.}          & \multicolumn{6}{c}{\textbf{Scene 20}~~~\textitgray{Speed: 0.849, Num: 837, Length 711.229 m}}          & \multicolumn{6}{c}{\textbf{Average}}                                                                \\
                              \midrule
\textitgray{Condition}                      & -               & \textitgray{Clone}        & \textitgray{Fog}         & \textitgray{Morning}          & \textitgray{Overcast}        & \textitgray{Rain}       & \textitgray{Sunset}      & \textbf{01 Avg.}     & \textbf{02 Avg.}     & \textbf{06 Avg.}             & \textbf{18 Avg.}      & \textbf{20 Avg.}     & \textbf{All Avg.}    \\
\midrule
DROID-SLAM \cite{droidslam}                   & \redtext{\textit{Required}}        & \first{3.5915}       & \first{5.0789}      & \first{3.7332}           & \first{3.8521}          & \first{3.7800}     & \second{4.9074}      & \second{1.1366}      & \first{0.0640}      & \first{0.0377}              & \second{2.2046}       & \first{4.1572}      & \first{1.5200}      \\
MASt3R-SLAM \cite{mast3r-slam}                  & \greentext{\textit{No Need}}         & \textitgray{TL}           & \textitgray{TL}          & \textitgray{TL}               & \textitgray{TL}              & \textitgray{TL}         & \textitgray{TL}          & /           & /           & /                   & /            & /           & /           \\
CUT3R \cite{cut3r}                        & \greentext{\textit{No Need}}         & 129.4984     & 76.9624     & 117.9483         & 114.5124        & 66.7003    & 116.5289    & 48.3618     & 20.1191     & 0.7724              & 17.1494      & 103.6918    & 38.0189     \\
Fast3R \cite{fast3r}                       & \greentext{\textit{No Need}}         & \textitgray{OOM}          & \textitgray{OOM}         & \textitgray{OOM}              & \textitgray{OOM}             & \textitgray{OOM}        & \textitgray{OOM}         & /           & /           & /                   & /            & /           & /           \\
VGGT \cite{vggt}                         & \greentext{\textit{No Need}}         & \textitgray{OOM}          & \textitgray{OOM}         & \textitgray{OOM}              & \textitgray{OOM}             & \textitgray{OOM}        & \textitgray{OOM}         & /           & /           & /                   & /            & /           & /           \\
\textbf{VGGT-Long (Ours)}              & \greentext{\textit{No Need}}         & \second{9.6553}       & \second{8.1851}      & \second{6.3447}           & \second{4.5642}          & \second{6.4990}     & \first{4.8500}      & \first{1.0490}      & \second{0.7026}      & \second{0.4377}              & \first{1.3966}       & \second{6.6830}      & \second{2.0538}     \\
\bottomrule
\end{tabular}

  }
  \caption{Camera Tracking Results (ATE RMSE [m] $\downarrow$) on the Virtual KITTI Dataset. Color definition: \first{first}, \second{second}.}
  \label{table:vkitti_ate}
\end{table*}

We evaluate VGGT-Long on three datasets, KITTI Dataset Odometry Track \cite{geiger2012kitti}, Waymo Open Dataset (v1.4.1) \cite{waymo}, and Virtual KITTI Dataset (v1.3.1) \cite{vkitti} to assess its performance in large-scale monocular 3D reconstruction. Our experiments are conducted on the computer with Ubuntu 22.04, and equipped with 12 Intel Xeon Gold 6128 3.40 GHz CPUs, 67GiB of RAM. In this paper, most experiments were conducted using an NVIDIA RTX 4090 GPU with 24 GiB of VRAM, while for experiments with a chunk size of 90 or larger, we used an NVIDIA L20 GPU with 48 GiB of VRAM.

\subsection{Metrics}

We report four different metrics. For tracking performance, following the recent literature, we use the Absolute Trajectory Error (ATE) metric \cite{ate} to evaluate the model's long-term consistency over extended sequences. For reconstruction metrics, we adopt the settings from VGGT \cite{vggt} and employ: 1) Accuracy: Euclidean distance from each predicted point to its nearest Ground Truth (GT) point; 2) Completeness: Euclidean distance from each GT point to its nearest predicted point; 3) Chamfer Distance: The average of the above two metrics. Due to the scale ambiguity inherent in monocular 3D methods, we first perform coarse alignment between the predicted poses and GT poses before calculating the reconstruction metrics. Subsequently, we refine the alignment using the Point-to-Point Iterative Closest Point (ICP) method \cite{icp} and then compute these metrics.

\subsection{Experiments Settings}

For the calculation of reconstruction performance metrics, to avoid interference from outlier point clouds, we uniformly retain points with confidence values greater than $0.75\times$ the average confidence for Transformer-based 3D methods. As for models like DROID-SLAM, which do not output confidence values, we retain points with depth values larger than $0.75\times$ the average inverse depth as the filtered point cloud.

Following the VGGT's configuration, we downsample input images to 518-pixel width while the hight maintaining aspect ratio during inference. Our inference pipeline first processes all images through VPR, then releases the VPR model's memory occupancy to allocate sufficient GPU memory resources for VGGT.

\subsection{CPU Memory Management Strategy}

To handle large-scale scenarios where CPU memory cannot accommodate all VGGT outputs, we implement a memory-efficient strategy: after processing each chunk, we store the results on disk. During subsequent alignment phases, we selectively load only relevant chunk pairs into CPU memory for SIM(3) computation, and immediately freeing the memory after calculation. This design offloads CPU memory pressure to disk storage, and will effectively prevents the operating system crashes, freezes, or other catastrophic failures caused by excessive CPU memory consumption. Upon completing computations, the model purges all intermediate results to minimize the storage overhead. The final outputs (colored point clouds and camera poses) employ stream writing techniques to bypass CPU memory constraints during large-scale reconstruction, effectively eliminating RAM pressure from handling massive point cloud datasets.

\subsection{KITTI, Waymo \& Virtual KITTI}

We begin with the KITTI odometry dataset, a classic benchmark for SLAM evaluation. Due to the long sequence lengths and outdoor settings, KITTI presents significant challenges for monocular reconstruction. Table~\ref{table:kitti_ate} reports the ATE across 11 sequences.

VGGT-Long outperforms the learning-based methods such as DROID-SLAM and DPVO. Unlike ORB-SLAM2 or LDSO, our method does not rely on calibrated camera intrinsics and still maintains competitive accuracy. Notably, VGGT-Long runs successfully on all sequences, while foundation models like Fast3R, CUT3R, and VGGT fail due to memory overflow. This verifies that our chunk-and-align framework effectively extends VGGT to long sequences without sacrificing scalability. When running MASt3R-SLAM on KITTI Odometry, we observe that tracking stalled after about 100 frames. New frames cannot be selected as keyframes, causing the mapping part to stop updating even though the process continues running. This phenomenon indicates that MASt3R-SLAM undergoes tracking lost. Tracking lost is defined when the system fails to register new frames for a prolonged period, leading to no further updates in the map and trajectory. Although Figure \ref{fig:loop} shows a better reconstruction correction and Table \ref{table:kitti_ate} demonstrates better tracking performance, we can still observe that there are still some parts of the central intersection that are not aligned. This is because there is no explicit loop closure constraint for the input image in this area.

To validate generalization, we test VGGT-Long on the Waymo Open Dataset, which features urban driving with high variability in scene appearance and traffic conditions. VGGT-Long achieves an average ATE of 1.996m across ten 200-frame segments. Compared to methods like CUT3R and MASt3R-SLAM, VGGT-Long yields significantly lower errors, as shown in Table~\ref{table:waymo_ate}. On segments with strong viewpoint diversity, VGGT-Long consistently produces more accurate and complete 3D reconstructions.

We further assess robustness under synthetic domain shifts using the Virtual KITTI dataset, which offers multiple weather and lighting conditions. As Table~\ref{table:vkitti_ate} shows, VGGT-Long maintains stable ATE across all tested conditions, including fog, rain, and sunset. Unlike DROID-SLAM, which occasionally fails to track, and CUT3R, which exhibits large drifts, our method remains robust without requiring retraining or domain adaptation.

In the experiments, the previous methods achieve lower tracking accuracies. The reasons for DROID-SLAM's poor performance on long outdoor sequences have been thoroughly discussed in \cite{gigaslam, dpv-slam}. MASt3R-SLAM works in short sequences but often fails in longer ones, a limitation tied to its reliance on MASt3R \cite{mast3r} for feature matching. In autonomous driving scenarios with less scene variation than that in indoor scenarios, like on a long straight road, the system may not generate new keyframes for extended periods. When it finally does, the significant distance from the last keyframe causes feature matching to fail, leading to tracking loss. As for CUT3R, it employs continuous state tokens to encode the entire 3D scene, which is analogous to NeRF \cite{mildenhall2021nerf} in novel view synthesis. Consequently, CUT3R faces similar challenges as NeRF in large-scale outdoor scenes. That is, a compact representation struggles to capture the vast geometric details of expansive large-scale outdoor environments.

\subsection{Loop Optimization Analysis \& Ablation Study}

Table \ref{table:runtime} reports the runtime of VGGT-Long on several KITTI sequences. All measurements exclude disk I/O time, as it can vary depending on OS background programs, disk bandwidth, and memory throughput. On our computer, each chunk takes approximately 25ms to load and 95ms to write, which is negligible given that each KITTI sequence contains only a few dozen chunks. Therefore, disk latency does not significantly impact overall runtime.

Our system achieves efficient chunk-wise processing (about 2.6–2.8s per chunk) and fast Sim(3) alignment (about 0.2s). We evaluate the effectiveness and efficiency of our loop optimization module in Fig.~\ref{fig:loop_convergence}. The proposed Sim(3)-based Levenberg-Marquardt solver converges within \textbf{3 iterations} on average and takes less than \textbf{15ms per step} (taking 0.4-1.3ms/iter in C++ or 3.5–13ms/iter in Python). These results demonstrate the practicality of our system in real-time or near real-time scenarios even on kilometer-scale sequences.

\begin{table}[h]
  \centering
 \resizebox{\linewidth}{!}{\textbf{}
\begin{tabular}{c|c|ccccc}
\toprule
\multicolumn{2}{c}{\textitgray{Chunk Size = 75}} & \textbf{VPR Model}  & \textbf{Chunk Process} & \textbf{Chunk Align} & \textbf{LM Opt. (C++)} & \textbf{LM Opt. (Python)} \\
\midrule
\textbf{Seq.}          & \textbf{Seq. Frames}         & \textitgray{Time / Frame} & \textitgray{Time / Chunk}    & \textitgray{Time / Iter}   & \textitgray{Time / Iter}     & \textitgray{Time / Iter}        \\
\midrule
00            & \textitgray{4542}                & 21.264 ms  & 2.811 s       & 0.284 s     & 1.249 ms      & 13.394 ms        \\
05            & \textitgray{2761}                & 17.023 ms  & 2.614 s       & 0.273 s     & 0.862 ms      & 8.446 ms         \\
06            & \textitgray{1101}                & 18.523 ms  & 2.728 s       & 0.278 s     & 0.436 ms      & 3.592 ms\\
\bottomrule
\end{tabular}
  }
  \caption{Runtime Analysis of each components of VGGT-Long. All the components are able to operate in almost real-time.}
  \label{table:runtime}
\end{table}

\begin{table}[h]
  \centering
 \resizebox{0.5\linewidth}{!}{\textbf{}

\begin{tabular}{ccc|cc}
\toprule
LC  & IRLS & Weight & 00    & 05    \\
\midrule
\ding{55}  & \ding{51}    & \ding{51}  & 58.69 & 36.01 \\
\ding{51} & \ding{55}     & \ding{51}  & 12.29 & 10.98 \\
\ding{51} & \ding{51}    & \ding{55}   & 11.28 & 10.13 \\
\midrule
\ding{51} & \ding{51}    & \ding{51}  & 8.67  & 8.31 \\
\bottomrule
\end{tabular}
  }
  \caption{Ablation study (ATE RMSE [m] $\downarrow$). Experiments demonstrate that loop closure significantly reduces cumulative errors, and the combination with all components yields optimal performance.}
  \label{table:Ablation}
\end{table}

Table~\ref{table:Ablation} presents an ablation study on KITTI sequences. “LC” indicates whether loop closure is enabled. “IRLS” denotes the use of iterative reweighted least squares in chunk alignment; disabling it results in a single-pass confidence-weighted alignment. “Weight” refers to whether confidence-based weighting is applied during alignment. To disable it, we normalize the VGGT confidence map to a uniform value across all points. Removing either the loop closure module or the IRLS weighting results in noticeable accuracy degradation. Specifically, removing loop closure increases the ATE to 58.69m on Seq. 00, while disabling IRLS leads to a 13\% drop in performance. The best performance is achieved when all components, loop closure, IRLS and confidence-weighted alignment are enabled.

\section{Conclusion}

In this work, we presented VGGT-Long, a simple yet effective framework that extends monocular RGB-only 3D reconstruction to long, unbounded video sequences using foundation models. Our method overcomes the GPU memory limitations of existing 3D vision models without requiring camera calibration. Through extensive experiments on KITTI, Waymo, and Virtual KITTI, we demonstrated that VGGT-Long achieves accurate and scalable 3D reconstruction across diverse real-world and synthetic environments. In the future, we will continue researching ways to improve the accuracy and consistency of 3D foundation models for long outdoor sequences.

{
    \small
    \bibliographystyle{ieeenat_fullname}
    \bibliography{main}
}

\clearpage

\begin{figure*}[t]
    \centering
    \includegraphics[width=\linewidth]{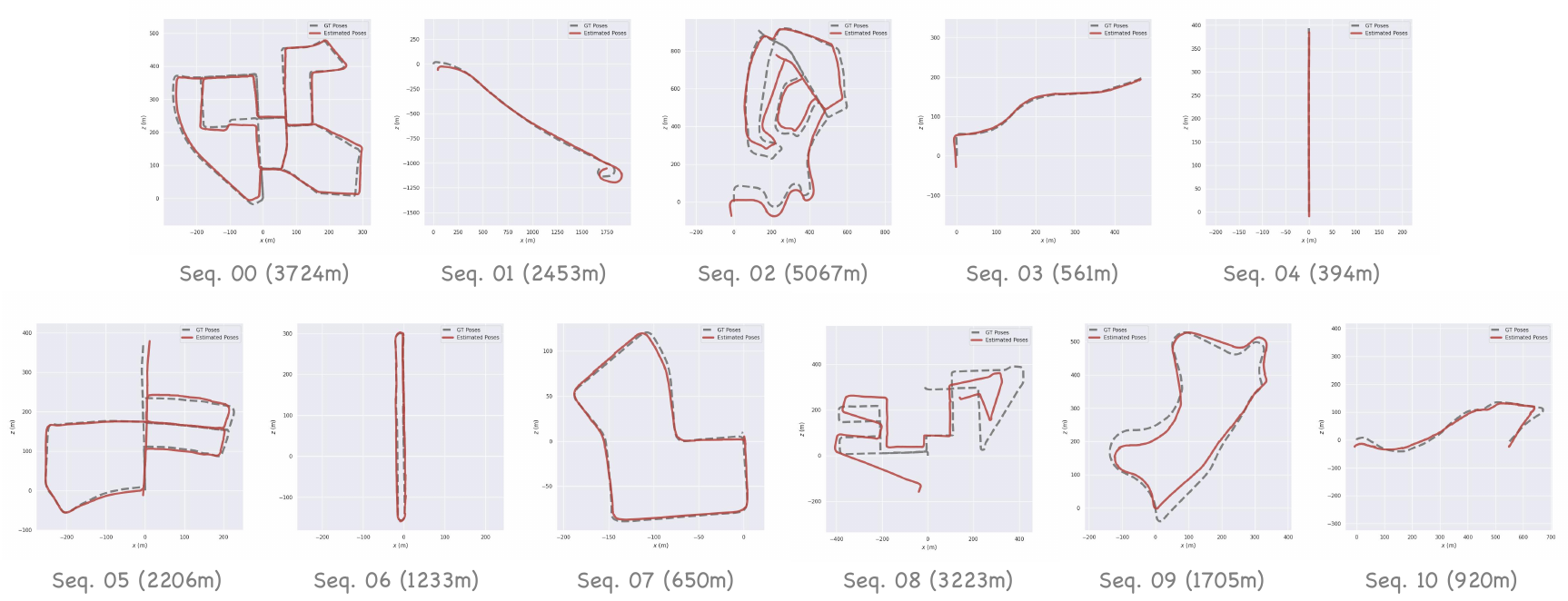}
    \caption{Trajectory visual results of KITTI Dataset. Although there is overlap in 3D space in Sequence 08, it often corresponds to different driving directions within the same lane. In such cases, the 2D RGB loop detection module fails to identify potential loops. Nevertheless, even under such circumstances, according to Table \ref{table:kitti_ate}, VGGT-Long still outperforms many existing methods in terms of numerical accuracy.}
    \label{fig:kitti_traj}
\end{figure*}

\begin{figure*}[t]
    \centering
    \includegraphics[width=\linewidth]{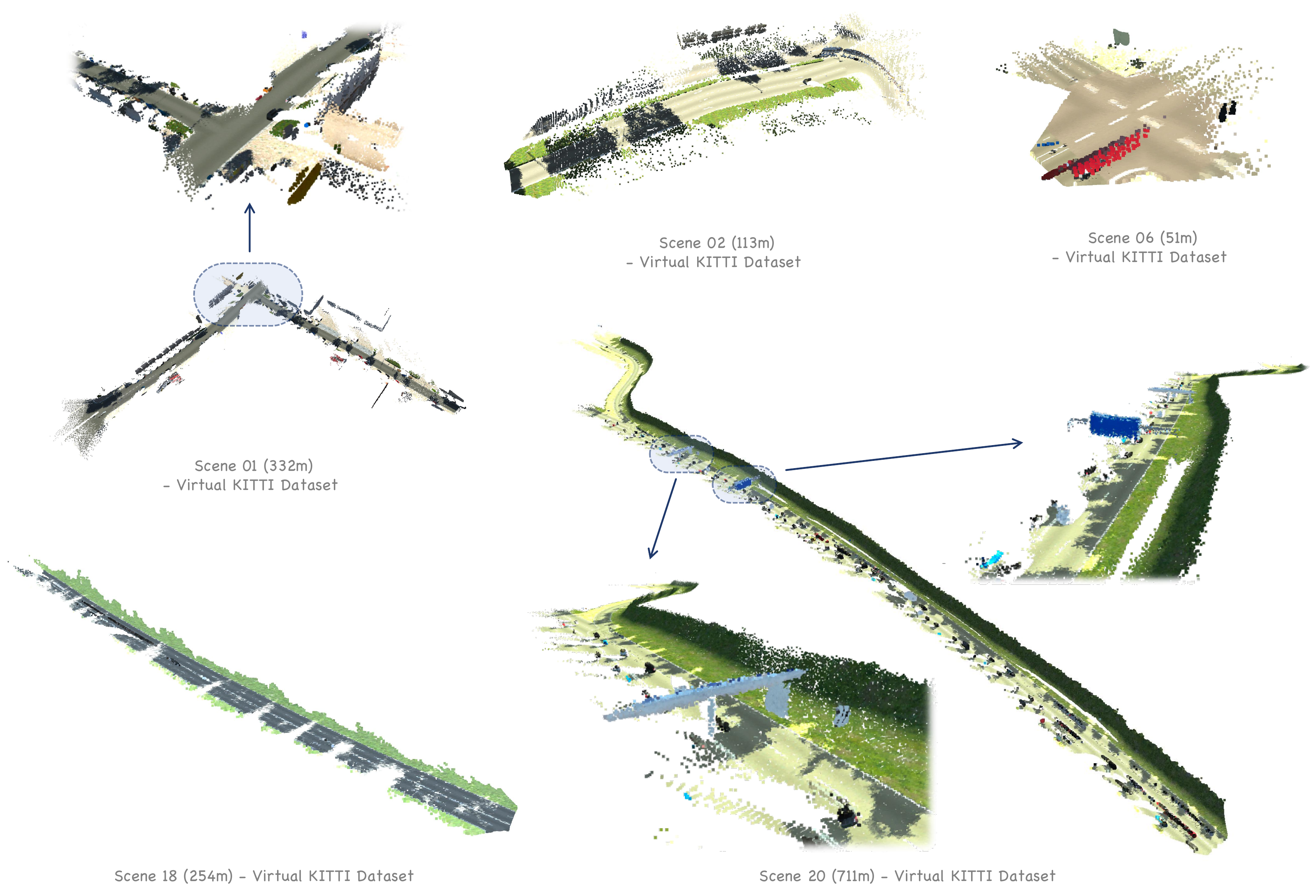}
    \caption{Visual results of Virtual KITTI Dataset.}
    \label{fig:vkitti}
\end{figure*}

\begin{figure*}[t]
    \centering
    \includegraphics[width=\linewidth]{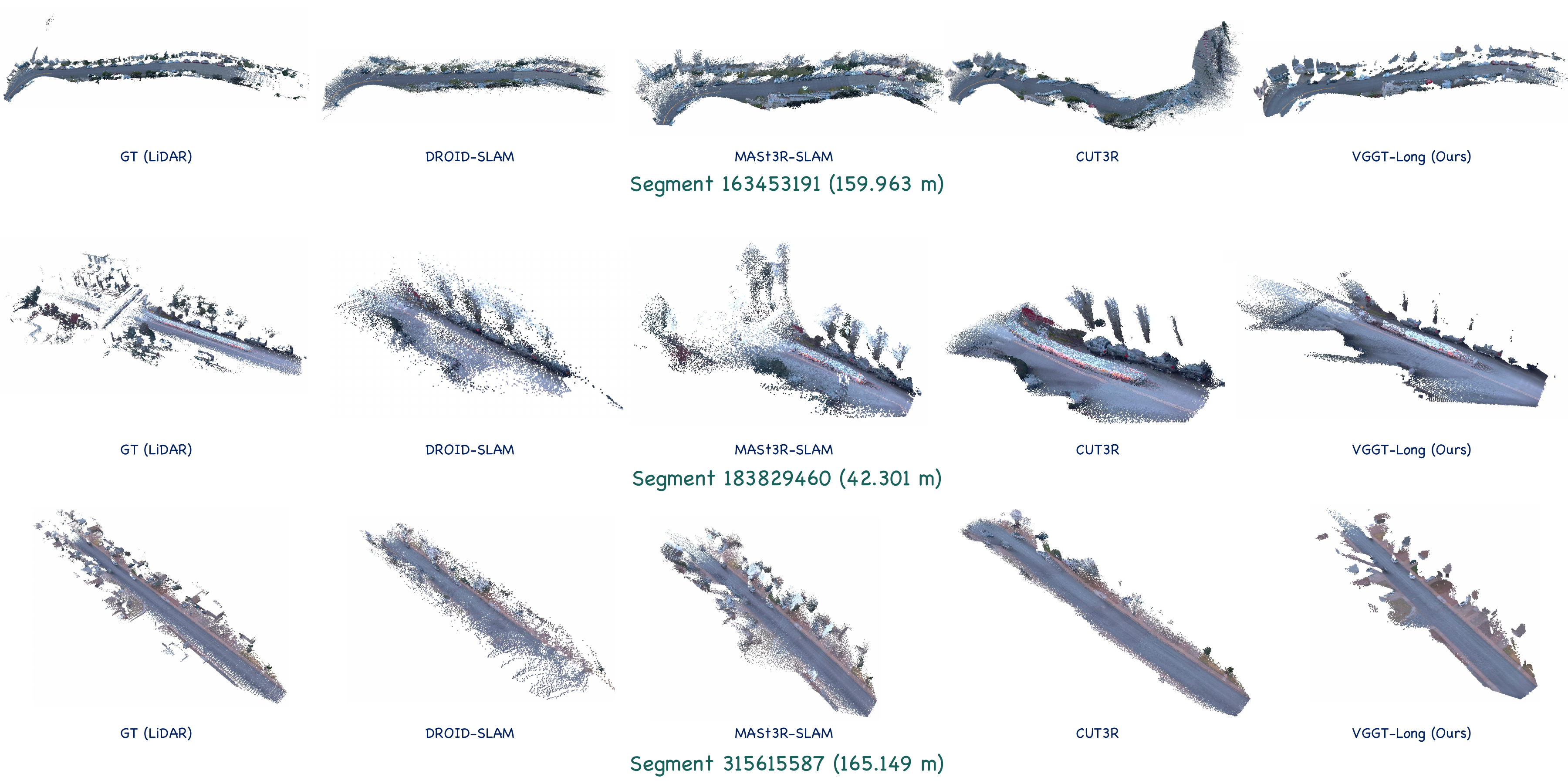}
    \caption{Visual comparison of Waymo Open Dataset. Part 1.}
    \label{fig:visual_cmp_1}
\end{figure*}

\begin{figure*}[t]
    \centering
    \includegraphics[width=\linewidth]{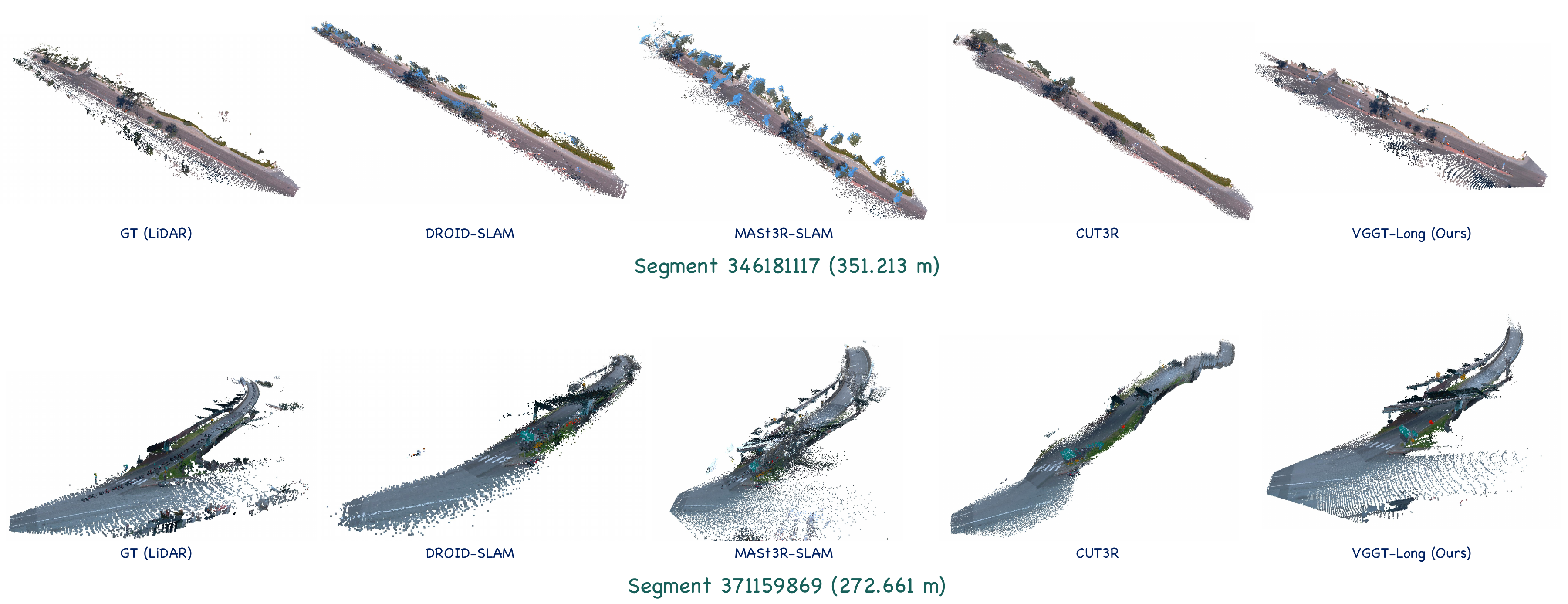}
    \caption{Visual comparison of Waymo Open Dataset. Part 2.}
    \label{fig:visual_cmp_2}
\end{figure*}

\begin{figure*}[t]
    \centering
    \includegraphics[width=\linewidth]{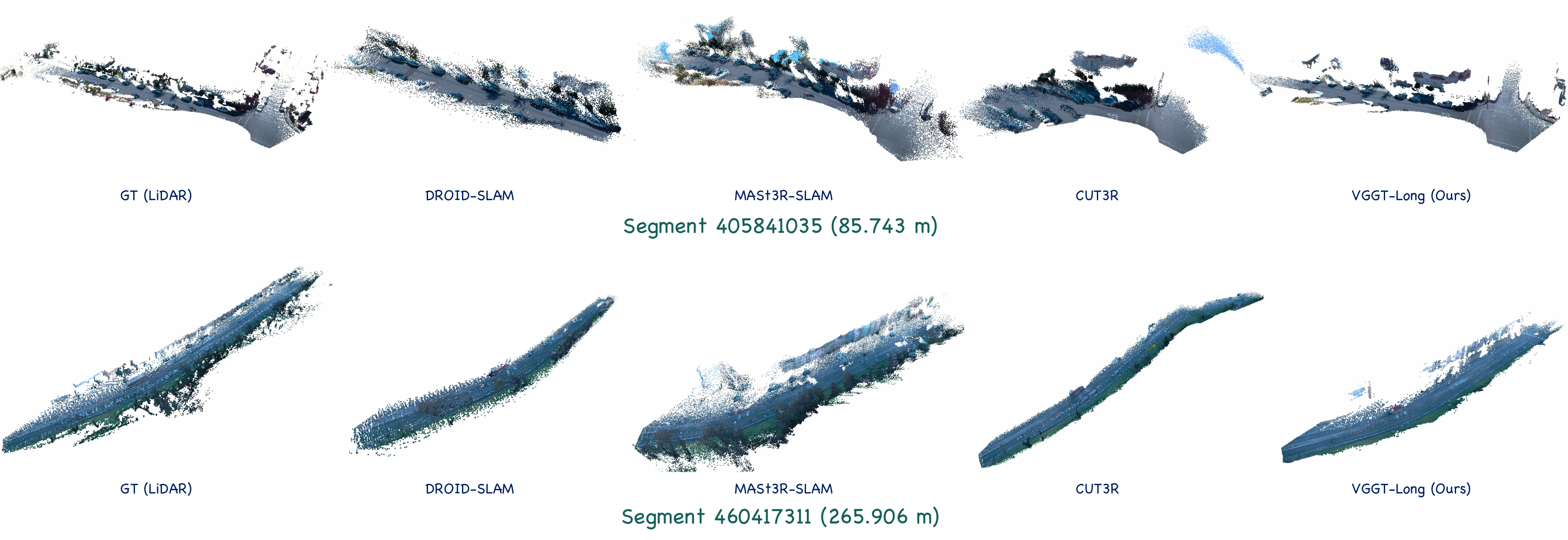}
    \caption{Visual comparison of Waymo Open Dataset. Part 3.}
    \label{fig:visual_cmp_3}
\end{figure*}

\begin{figure*}[!t]
    \centering
    \includegraphics[width=\linewidth]{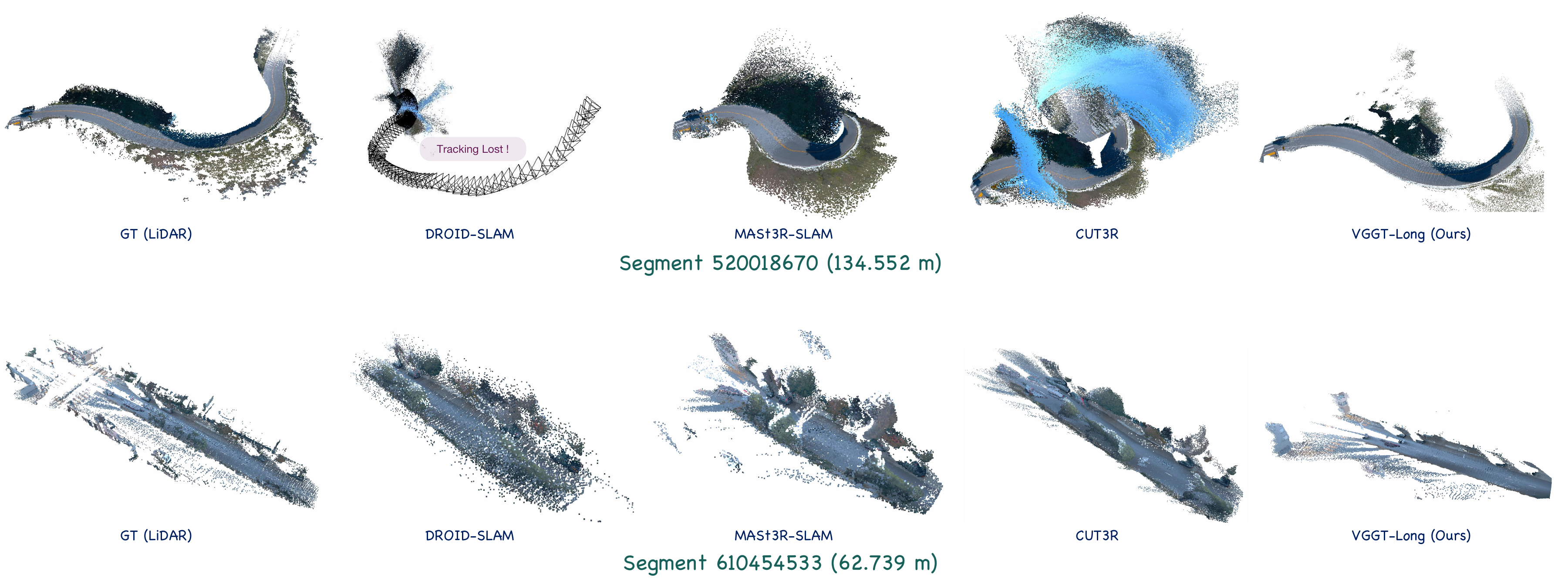}
    \caption{Visual comparison of Waymo Open Dataset. Part 4.}
    \label{fig:visual_cmp_4}
\end{figure*}



\end{document}